\documentclass[conference]{IEEEtran}
\usepackage{times}

\usepackage[numbers]{natbib}
\usepackage{multicol}
\usepackage[bookmarks=true]{hyperref}
\usepackage{graphicx} 
\usepackage{hyperref}
\usepackage{wrapfig}
\usepackage{xcolor}
\usepackage{booktabs}
\usepackage{fancyhdr}

\begin{document}

\title{PlantTrack: Task-Driven Plant Keypoint Tracking with Zero-Shot Sim2Real Transfer}



%
\author{\authorblockN{Samhita Marri,
Arun N. Sivakumar,
Naveen K. Uppalapati and
Girish Chowdhary}\\
\authorblockA{ Field Robotics Engineering and Science Hub (FRESH)\\
University of Illinois at Urbana-Champaign (UIUC), IL, USA\\
Email: (marri2, av7, uppalap2, girishc)@illinois.edu}%
\thanks{This paper was supported in part by the National Science Foundation under Grant No. DBI-2019674. }}

\maketitle

\thispagestyle{fancy}

\begin{abstract}

Tracking plant features is crucial for various agricultural tasks like phenotyping, pruning, or harvesting, but the unstructured, cluttered, and deformable nature of plant environments makes it a challenging task. In this context, the recent advancements in foundational models show promise in addressing this challenge. In our work, we propose PlantTrack where we utilize DINOv2 which provides high-dimensional features, and train a keypoint heatmap predictor network to identify the locations of semantic features such as fruits and leaves which are then used as prompts for point tracking across video frames using TAPIR. We show that with as few as 20 synthetic images for training the keypoint predictor, we achieve zero-shot Sim2Real transfer, enabling effective tracking of plant features in real environments. 

\end{abstract}

\IEEEpeerreviewmaketitle

\section{Introduction}

Our global population is projected to reach around 10 billion by the end of 2025 according to World Population Prospects 2019 highlights \cite{United2019}. The growing labor shortage in the agricultural sector makes it difficult to meet the food demands. These factors in addition to climate change, pose a significant challenge and require us to develop sustainable agricultural practices. Deploying autonomous robotic solutions not only tackles the labor shortage but also helps in phenotyping in developing resilient crops which is primarily done manually.

Tracking plant features is crucial for various tasks in agricultural robotics such as phenotyping, pruning, and harvesting. The complex nature of the plant environments due to their cluttered physical structure, deformability, and external dynamic factors such as wind, and lighting, make it not so straightforward. For instance, the position of the leaf or fruit can change over time due to external wind forces, while the robot passes through crop rows, or interacts with the plant.


Traditional computer vision deploys techniques such as optical flow \cite{lucas1981iterative} or template-based matching \cite{briechle2001template} to track the features across different images. However, these methods require careful handcrafting of features, making them less robust to variations in lighting or occlusion. The advent of neural networks made it possible to extract features, for computer vision tasks such as image classification, bounding box detection, or segmentation. However, these methods require large amounts of data and labeling, which is labor-intensive and not scalable to deploy to different plant tasks or plant varieties. Additionally, labeling for keypoints detection tasks is not as straightforward as labeling for segmentation or object recognition tasks.

\begin{figure}[!t]
\centering
\includegraphics[width=0.48\textwidth]{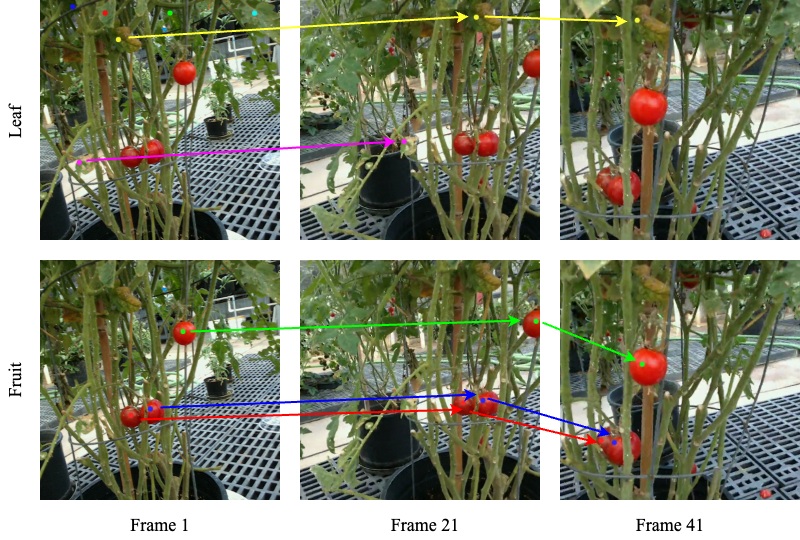}
\caption{Predicted keypoint tracking using TAPIR at various frames where (top) leaves and (bottom) fruits predicted in the first frame are given as prompt.}
\label{fig:tapir}
\vspace{-0.8cm}
\end{figure}

\begin{figure*}[!h]
\includegraphics[width=\textwidth]{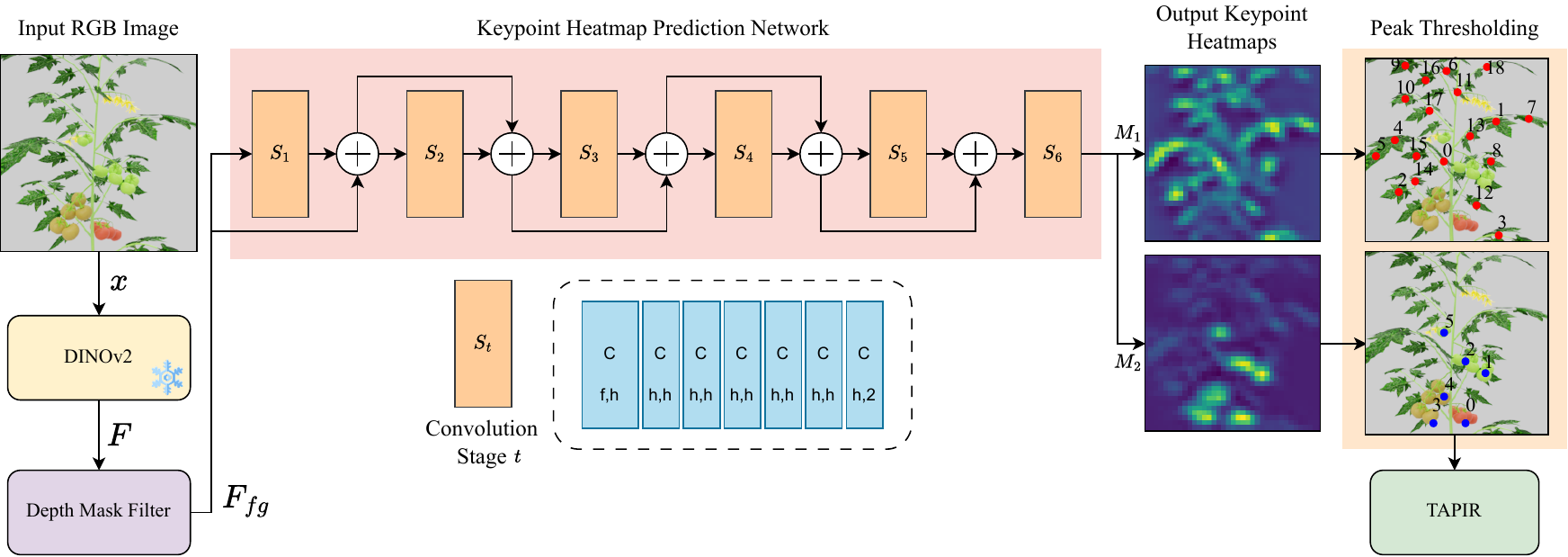}
\caption{PlantTrack: First we extract high-dimensional features from RGB image using DINOv2. Then, we apply the depth mask to focus only on the foreground features. These foreground features are passed through the keypoint heatmap prediction network which is a convolution-based architecture. The input and output channels for each convolution block $C$ are indicated by $f, h, 2$ values. Once the heatmap prediction network is trained, peak pixel locations across the image are predicted and are used as prompts to perform tracking across video frames using TAPIR.}
\label{fig:pipeline}
\end{figure*}

Therefore in this work, we propose PlantTrack as shown in Figure \ref{fig:pipeline}, which includes the use of synthetic data requiring minimal effort in obtaining the ground truth labels and eliminating the ambiguity in the labeling process. Further motivated by the latest advancements in foundational models, particularly the DINOv2 model \cite{oquab2023dinov2}, which predicts robust visual features from an image, we utilize it as a high-dimensional feature extractor to train our task-specific features. This approach also minimizes the training effort required to extract robust features from scratch. 

We also propose a keypoint heatmap prediction network trained using synthetic image data by obtaining corresponding high-dimensional features using DINOv2. The ground truth labels of keypoint heatmaps for this network are also generated synthetically. Once trained, the peak locations in the outputs of the keypoint predictor network are used to perform keypoint tracking using TAPIR \cite{doersch2023tapir}. This tracking is done in an online manner, where the keypoints are provided only in the first frame.

Furthermore, we demonstrate that the trained network can be deployed not only on unseen synthetic plant images but also on real plants, achieving zero-shot Sim2Real transfer for tracking leaves and fruits.

\section{Proposed Methodology}

Our proposed approach, PlantTrack, as shown in Figure \ref{fig:pipeline} first involves extracting the high-dimensional features using the DINOv2 \cite{oquab2023dinov2}. These high-dimensional features serve as input for training our keypoint regressor to predict heatmaps. Finally, these keypoints are used as prompts for tracking over time in an online manner using TAPIR \cite{doersch2023tapir}. Note that the model weights of both DINOv2 and TAPIR remain frozen and are not retrained in our approach. Instead, we focus on learning the keypoints, explained in the remainder of this section.

\begin{figure}[!h]
\includegraphics[width=0.45\textwidth]{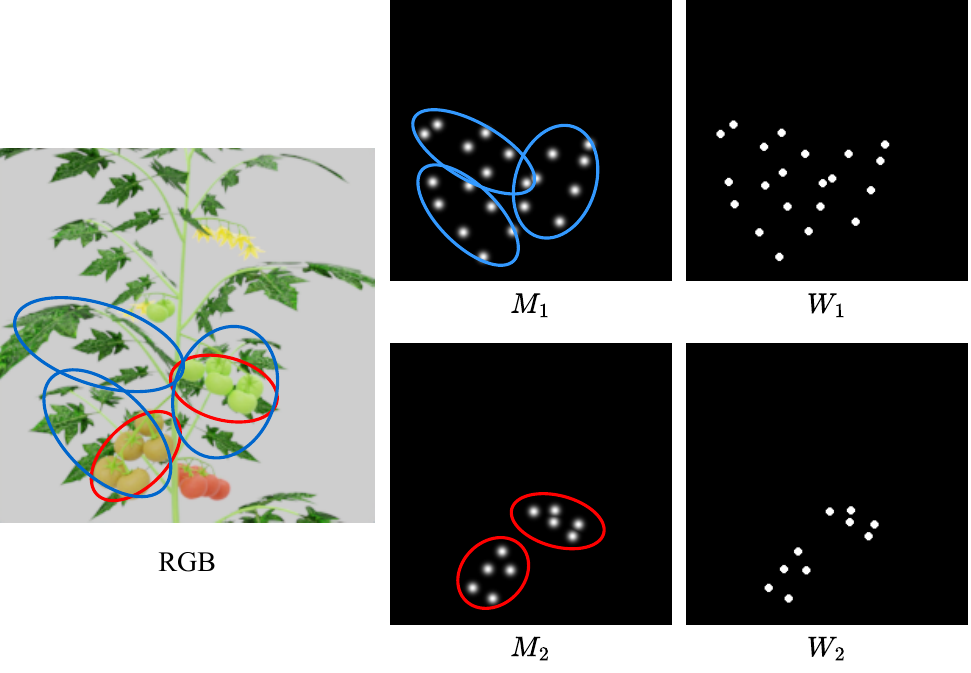}
\caption{Sample data collected in Blender using transformed global coordinates of the center of fruits and leaves, in (left) current camera view, (center) the heatmaps and (right) binary masks are obtained. }
\label{fig:data_sample}
\end{figure}

\subsection{Data Collection}

The input to the keypoint regressor is a high-dimensional feature image $F$ of dimension $(w, h, f)$, obtained from DINOv2 for an RGB image input $x$. Here, $f=384$ since we utilized the smallest DINOv2 model. The regressor then predicts two heatmaps $M_1, M_2$, each with dimensions $(w, h, 1)$, representing the centers of leaves and fruits respectively. The sample heatmap corresponding to an RGB image is shown in Figure \ref{fig:data_sample}. 

Leveraging a 3D model of a cherry tomato plant in Blender, we collect RGB and depth images using the \href{https://github.com/DIYer22/bpycv/tree/master}{BPYCV} module. We then render corresponding keypoints heatmaps by transforming the global coordinates of fruit and leaf centers into local coordinates. This approach significantly reduces labeling effort while ensuring labeling consistency. It is worth noting that not all parts of the plant require labeling, as DINOv2 predicts consistent features for similar parts within the image, which is exploited by the keypoint predictor network.

\begin{figure*}[h]
\centering
\includegraphics[width=0.95\textwidth]{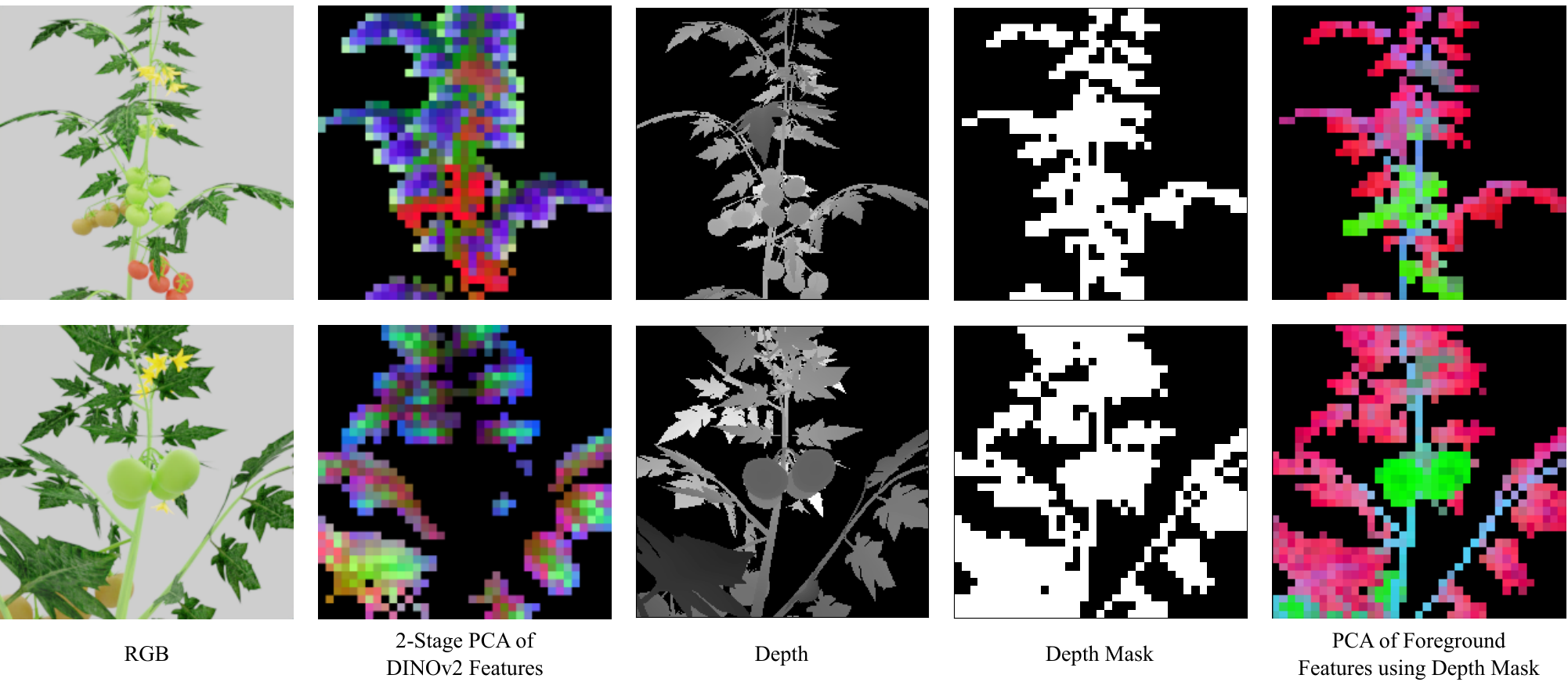}
\caption{Depth Mask Filter: To focus only on the foreground features 2-stage PCA is unreliable as shown in the bottom row, where the features of fruits are missing. It requires fine-tuning of the threshold parameter which is a tedious process to scale. Therefore, we implement filtering out the background using a depth mask, and as seen in the last column, all the features remain intact. Note that we use all the DINOv2 predicted features in the foreground after filtering with a depth mask and PCA is shown here just for visualization purposes. Also, the low resolution in columns 2, 4, and 5 is due to the downscaling of the input image by a factor of 14 in DINOv2.}
\label{fig:depth_mask}
\end{figure*}

Furthermore, we apply a depth mask over the DINOv2 output $F$ to obtain high-dimensional features only in the foreground $F_{fg}$, helping the keypoint regressor to focus only on the foreground part of the image. While performing principle component analysis (PCA) on the high-dimensional features as shown in the original DINOv2 work helps in distinguishing foreground from background, it requires fine-tuning the threshold parameter which is not robust as shown in Figure \ref{fig:depth_mask}.

Additionally, binary masks $W_1, W_2$ for each keypoint heatmaps are also obtained which helps in not penalizing the true positives during the training process as shown in \cite{cao2017realtime}.

\subsection{Training}

Coming to the training part of the keypoint regressor, we use a convolution-based architecture similar to the heatmap or confidence map prediction in OpenPose \cite{cao2017realtime}. The multi-stage convolution with cascading outputs from each stage to the next stage is known for better feature propagation and improved heatmap predictions \cite{cao2017realtime}. For a data sample $i$, $[F_{fg}^i, {M_1}_{gt}^i, {M_2}_{gt}^i, {W_1}^i, {W_2}^i]$, in total dataset of size $L$, and predictions at stage $t$, $[{M_1}_{t, p}^i, {M_2}_{t, p}^i]$, the MSE loss function in Equation \ref{eq1} as shown below is implemented.

\begin{equation}\label{eq1}
    L_t = \sum_i^N W_1^i||{M_1}_{gt}^i - {M_1}_{t, p}^i||^2 + 
    \sum_i^N W_2^i||{M_2}_{gt}^i - {M_2}_{t, p}^i||^2
\end{equation}

The final loss function for $T$ stages is shown in Equation \ref{eq2} which is utilized to update the weights of the keypoint heatmap predictor network.

\begin{equation}\label{eq2}
    L = \sum_{t=1}^T L_t
\end{equation}

\section{Results and Discussion}

\begin{figure*}[!h]
\centering
\includegraphics[width=0.95\textwidth]{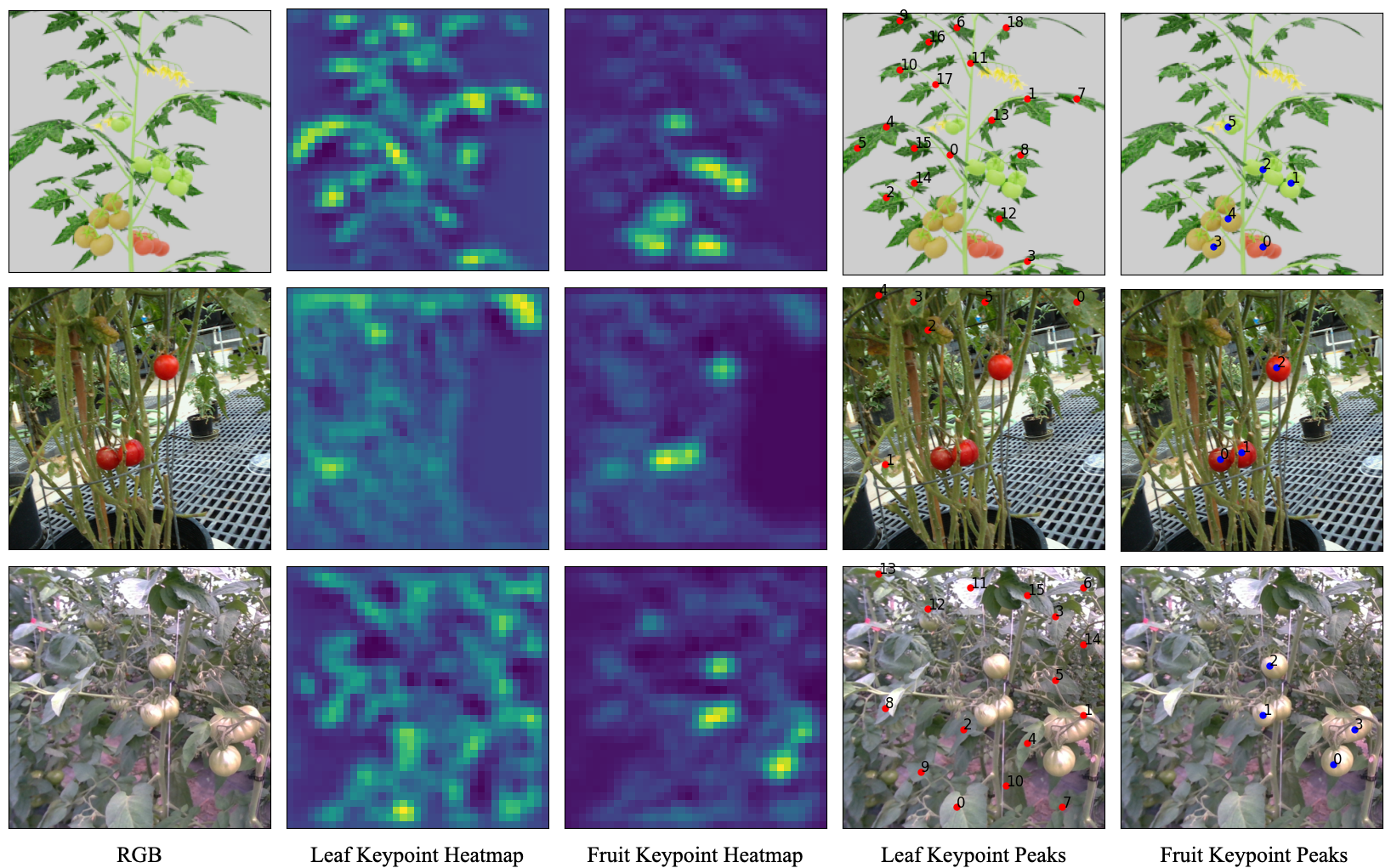}
\caption{Keypoint Heatmap Prediction Inference: Testing on (top row) unseen synthetic plant, and zero-shot detection on (middle row) real plant and (bottom row) real plant with a cluttered background. The leaf heatmap predictions are shown in column $2$ and leaf and fruit heatmap predictions are shown in column $4$. After normalizing and collecting peaks $>0.6$, the leaf (indicated in red) and fruit (indicated in blue) keypoints in columns $4$ and $5$ respectively.}
\label{fig:results}
\end{figure*}

We present results of heatmap predictions on test synthetic data collected in Blender unseen during training as well as zero-shot detection on real plants. Using the pipeline shown in Figure \ref{fig:pipeline} to predict the keypoint heatmap, we obtain the results shown in Figure \ref{fig:results}. With just $20$ images of training data, we show qualitative results of the heatmap predictions of not only the unseen synthetic data in Blender but also on a completely new environment in Figure \ref{fig:results}. We utilize ZoeDepth \cite{bhat2023zoedepth} to extract the depth image for depth filtering to test on images of real plants but depth cameras can also be deployed for practical purposes. We further use the predicted keypoints as pixel prompts to TAPIR \cite{doersch2023tapir}, which is a point tracker in a video, as shown in Figure \ref{fig:tapir}. This shows that we can exploit the foundational models in tracking the features of plants that are task-specific as shown in this work, particularly for fruits and leaves. 

While the results look promising in performing zero-shot keypoint detection on real plants, we would like to acknowledge that our work requires further improvements. First, filtering heatmap predictions requires a threshold and neighborhood size to look for peaks, which can increase the number of false positives or remove the true positives based on these parameters as seen in Figure \ref{fig:results} where not all keypoints are being detected after filtering. Second, to predict more keypoints for fruit such as its stem location or bottom in addition to its center requires higher input resolution and geometric constraints to reason about keypoint association within each part. Therefore, introducing geometry via depth, adding upsampling layers, and interpolating the features can help address the issue of resolution and also reason about occlusion. Further, TAPIR loses track as also seen in the leaf tracking in Figure \ref{fig:tapir} due to severe appearance changes in some of the keypoint (cyan) regions. Predicting prompts using our keypoint predictor network after certain frames along with the use of camera pose can help alleviate this issue.

\section{Conclusion} \label{sec:conclusion}

In this work, we propose our PlantTrack approach that utilizes DINOv2's high-dimensional features to learn keypoints, enabling the extraction of task-specific locations within images. We demonstrate the effectiveness of our approach by showcasing Sim2Real transfer of keypoint predictions and performing keypoint tracking using TAPIR on both leaves and fruits. Our findings highlight that foundational models are powerful in capturing robust features and can be enhanced through training with synthetic data for extracting task-specific features. Notably, our approach allows for deployment in real plant environments in a zero-shot manner. Moving forward, our research will focus on making further improvements to deploy for various plant manipulation tasks.

\section{Acknowledgement}

This work was supported by the Center for Research on Programmable Plant Systems (CROPPS) through the National Science Foundation under Grant No. DBI-2019674.

\bibliographystyle{IEEEtran}
\bibliography{references}

\end{document}